%% file: neurips_2022.tex
\documentclass{article}

\PassOptionsToPackage{numbers, compress}{natbib}

    \usepackage[preprint]{neurips_2022}

\usepackage[utf8]{inputenc} 
\usepackage[T1]{fontenc}    
\usepackage{hyperref}       
\usepackage{url}            
\usepackage{booktabs}       
\usepackage{amsfonts}       
\usepackage{nicefrac}       
\usepackage{microtype}      
\usepackage{xcolor}         
\usepackage{graphicx}
\usepackage{wrapfig}
\usepackage{caption}
\usepackage{multirow}
\usepackage[ruled,vlined, linesnumbered]{algorithm2e}
\usepackage{stmaryrd}
\usepackage{xr}
\usepackage{tabularx}
\usepackage{amsmath}
\usepackage{epigraph}
\newcommand{\myparagraph}[1]{\vspace{0pt}\paragraph{#1}}
\title{End-to-end View Synthesis via NeRF Attention}

\author{
  Zelin Zhao \\
  The Chinese University of Hong Kong\\
  \texttt{sjtuytc@gmail.com} \\
   \And
   Jiaya Jia \\
   The Chinese University of Hong Kong \& SmartMore \\
   \texttt{leojia@cse.cuhk.edu.hk} \\
}

\begin{document}

\maketitle

\input{text/0_abstract}
\input{text/1_introduction}
\input{text/2_related_work}
\input{text/3_method}
\input{text/4_experiments}
\input{text/5_conclusion}
\clearpage
\newpage

{
\small
\bibliography{neurips_2022}
\bibliographystyle{splncs04}
}
\end{document}

%% file: text/0_abstract.tex
\begin{abstract}
In this paper, we present a simple seq2seq formulation for view synthesis where we take a set of ray points as input and output colors corresponding to the rays. Directly applying a standard transformer on this seq2seq formulation has two limitations. First, the standard attention cannot successfully fit the volumetric rendering procedure, and therefore high-frequency components are missing in the synthesized views. Second, applying global attention to all rays and pixels is extremely inefficient. Inspired by the neural radiance field (NeRF), we propose the NeRF attention (NeRFA) to address the above problems. On the one hand, NeRFA considers the volumetric rendering equation as a soft feature modulation procedure. In this way, the feature modulation enhances the transformers with the NeRF-like inductive bias. On the other hand, NeRFA performs multi-stage attention to reduce the computational overhead. Furthermore, the NeRFA model adopts the ray and pixel transformers to learn the interactions between rays and pixels. NeRFA demonstrates superior performance over NeRF and NerFormer on four datasets: DeepVoxels, Blender, LLFF, and CO3D. Besides, NeRFA establishes a new state-of-the-art under two settings: the single-scene view synthesis and the category-centric novel view synthesis.
\end{abstract}

%% file: text/1_introduction.tex
\section{Introduction}
\begin{figure}[t]
  \centering
  \includegraphics[width=\textwidth]{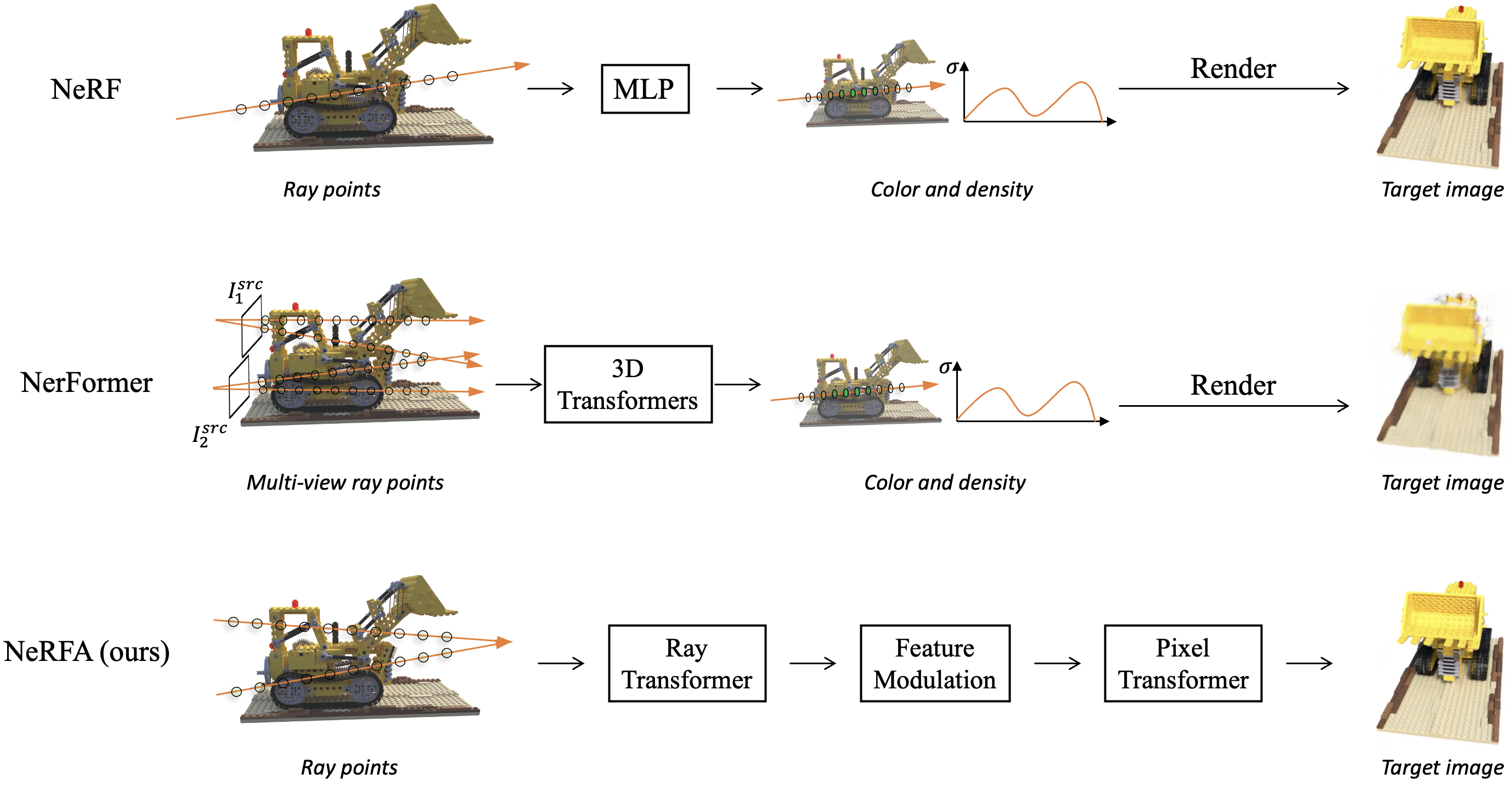}
  \caption{Essential difference between NeRF~\cite{nerf}, NerFormer~\cite{nerformer} and NeRFA. NeRF first predicts the color and densities of the ray points in 3D space, then renders them into a 2D image. NerFormer replaces the MLP with 3D transformers and builds color and densities from multiple source views. We instead address the view synthesis problem in a seq2seq manner where the NeRFA model transfers ray points to pixel colors in the target image.}
  \label{Fig:teaserFig}
\end{figure}

The view synthesis task is to generate a novel image from an arbitrary viewpoint~\cite{nerf, viewSynthesis}. We consider a simple seq2seq formulation: the input includes ray points (points sampled on rays), and the output is the colors contributed by the corresponding rays in the image plane (as shown in  Figure~\ref{Fig:teaserFig}). This formation does not require rendering and does not require the 3D model of objects. With this simple formulation, we are heading towards a uniform end-to-end machine learning model~\cite{foundationModels}.

One possible solution for this problem formulation is the transformer~\cite{transformer} which has become the dominant neural model in many computer vision and natural language processing tasks~\cite{transformer, SwinTransformer,lxmert,DETR}. We first try the standard transformer under the seq2seq formulation. It is found that the vanilla transformer yields poor performance (refer to Figure~\ref{Fig:ablation_result}). We hypothesize the following reasons for this fact. On the one hand, the standard global attention lacks the inductive bias of the volumetric rendering process, and the resulting images tend to be over-smoothed. On the other hand, global attention is computationally heavy, so the training is extremely inefficient.

To address these issues, we propose the NeRF attention (NeRFA), inspired by neural radiance fields (NeRF~\cite{nerf}). We turn the volumetric rendering process into a feature modulation procedure to empower the transformer with the latent rendering ability. Unlike NeRF, we do not explicitly predict the mid-way colors and densities and instead conduct latent rendering via the attention mechanism. Therefore, NeRFA preserves the attention mechanism's representation potential while exploiting the inductive bias inspired by the volumetric rendering equation. Besides, the NeRFA computes ray attention and pixel attention separately to save computational costs~\cite{SwinTransformer}.

Importantly, NeRFA is the first to provide a full-attention solution to the novel view synthesis task. The most related work, NerFormer~\cite{nerformer} is published accompanied by the CO3D dataset. As shown in Figure~\ref{Fig:teaserFig}, NerFormer replaces the MLP in NeRF with the transformers and consumes multi-view input to generate color and density predictions. It does not eliminate the volumetric rendering and is of different model architecture from NeRFA.

NeRFA generates photo-realistic images without explicit 3D object models. It inherits the power of NeRF by learning latent volumetric representation in continuous 3D space. In addition, NeRFA adopts transformers to learn interaction between ray points and builds connections regarding colors of surrounding pixels, which helps generate sharper regions in complex scenes.

We conduct extensive experiments to validate the effectiveness of the NeRFA model. Visual comparison between NeRF and NeRFA shows that NeRFA produces better fine-grained details of objects. The quantitative results also manifest that the NeRFA model consistently outperforms NeRF~\cite{nerf} and NerFormer~\cite{nerformer} on Blender~\cite{nerf}, LLFF~\cite{localLightFieldFusion}, and the DeepVoxels~\cite{deepvoxels_sitzmann} datasets for the single-scene view synthesis. Furthermore, NeRFA also surpasses the NeRF~\cite{nerf} and NerFormer~\cite{nerformer} on the CO3D~\cite{nerformer} dataset in the category-centric novel view synthesis.

%% file: text/2_related_work.tex
\section{Related work}
\myparagraph{View synthesis} Early graphics work, such as the Lumigraph~\cite{Lumigraph}, reconstructs novel views via reconstructed light field. Later work operates on mesh-based representation to assign color~\cite{SurfaceLight} or add texture~\cite{LargeScaleTexturing,6dpose,rapl} to 3D objects. In \cite{viewSynthesis}, a CNN predicts the appearance flow from the input to the output view. \cite{wiles2020synsin} presents an end-to-end point cloud-based framework Synsin for single-image view synthesis. This setting is different from our multi-frame view synthesis task. 

Another line of work leverages the volumetric representation to synthesize high-quality views from observed images. DeepView~\cite{DeepView} and the follow-up work~\cite{PushingBoundariesOfView,localLightFieldFusion} synthesize photo-realistic images via multi-plane images (MPIs). \cite{LearningMultiViewStereo, soft3dreconstruction, MultiViewSupervision} use learnable volumetric rendering to render views. They can hardly generate photo-realistic natural images.

\myparagraph{Neural radiance field} The neural radiance field (NeRF~\cite{nerf}) provides a simple yet effective solution to the view synthesis problem. The NerFormer model is presented along with the CO3D dataset in \cite{nerformer}. It replaces the MLP module with a transformer encoder and does not change the ray march module. We compare the NerFormer thoroughly in the experiments. In \cite{geoViewSynthesis}, a transformer architecture for the single-view synthesis task was proposed. Their setting is different from ours. The VQGAN~\cite{vqganTamingTransformer} structure is adopted to address the ambiguity issue in single views. Neural sparse voxel fields (NSVF~\cite{NSVF}) use voxel octrees to boost inference in learning scene representations. Other recent work speeds up training~\cite{directVoxelOptimization} and inference~\cite{autoint,fastnerf,kilonerf} of NeRF. The work of \cite{nerfies,dnerf,dynamicNeRF,pva} reconstructs a non-rigid deformable scene from captured images. Further, GRAF~\cite{graf} and Giraffe~\cite{giraffe} propose generative networks based on NeRF to synthesize new images. IBRNet~\cite{ibrnet} uses a ray transformer to summarize densities from neighborhood views. Another important application of NeRF is scene reconstruction~\cite{neus,VolumeRO}. 


\myparagraph{Neural rendering} Researchers made extensive efforts to pursue learnable photo-realistic rendering models. RenderNet~\cite{rendernet} is a convolutional neural network learned to perform projection and shading. The visual object networks~\cite{von} learn to generate realistic images from learned disentangled 3D representations. Textured neural avatars~\cite{texturedNeuralAvatars} generate full-body rendering of a person via image-to-image deep networks. In \cite{deferredNeuralRendering}, neural textures improve the rendering quality. These methods rely on the 3D models of objects and therefore cannot be applied to general natural images directly. The concurrent work ViewFormer~\cite{viewformer} proposes a NeRF-free neural rendering, which addresses the view synthesis problem in 2D space, so their synthesis quality is limited.

\myparagraph{Attention mechanisms and transformers} The attention mechanism was proposed in machine translation~\cite{transformer}. Transformers then demonstrate power in many different domains, such as cross-modal reasoning~\cite{lxmert, ProTo}, large-scale unsupervised pretraining~\cite{bert, sun2019videobert, xlnet}, and multi-task representation learning~\cite{languageMultiTask, sentimentAnalysis, p3aformer}. Recently, transformers become competitive models in fundamental vision tasks, such as image classification~\cite{visionTransformer, REST}, object detection~\cite{DETR}, and segmentation~\cite{SwinTransformer, container}. Transformers demonstrate high generalization~\cite{howDoViTWork} ability and serve as foundation models~\cite{foundationModels}.

%% file: text/3_method.tex
\section{Problem formation and approach}
In Sec.~\ref{sec:problem_form}, we present a simple seq2seq formulation for the view synthesis problem. We then introduce the vanilla attention model in Sec.~\ref{sec:vanilla_atten}. Subsequently, we introduce the NeRF attention in Sec.~\ref{sec:nerfa}, which addresses key limitations of the vanilla attention model.

\subsection{End-to-end view synthesis}
\label{sec:problem_form}
As shown in Figure~\ref{Fig:teaserFig}, we uniformly sample sequences of points on rays in the 3D space, and we call them ray points (they can also be called query points~\cite{nerf}). The input to the model is a set of ray points in the 3D space, and the output is a set of pixel colors corresponding to those ray points. Formally, the input ray points are represented by a matrix $\mathbf{P}$ of shape $N_p\times N_r\times6$ where $N_r$ is the number of ray points corresponding to one ray, and $N_p$ is the number of rays sampled in one image. Since each ray corresponds to a projected pixel in the image plane, $N_p$ is also the number of sampled pixels. Here one ray point is represented by a 6D tuple $(x, y, z, d_x, d_y, d_z)$ where $(x, y, z)$ is the 3D location and $(d_x, d_y, d_z)$ is the unit vector representing view direction. The output is the RGB color $\mathbf{C} \in \mathbb{R}^{N_p \times 3}$. Given the ground-true RGB color $\mathbf{C}^{gt} \in \mathbb{R}^{N_p \times 3}$, we can train the model via a simple L2 loss of $\mathcal{L} = \left\|\mathbf{C}^{gt}-\mathbf{C}\right\|_{2}^{2}$. This simple end-to-end framework eliminates the explicit rendering procedure and the hierarchical volume sampling proposed in NeRF~\cite{nerf}.

\begin{figure}[t]
  \centering
  \includegraphics[width=\textwidth]{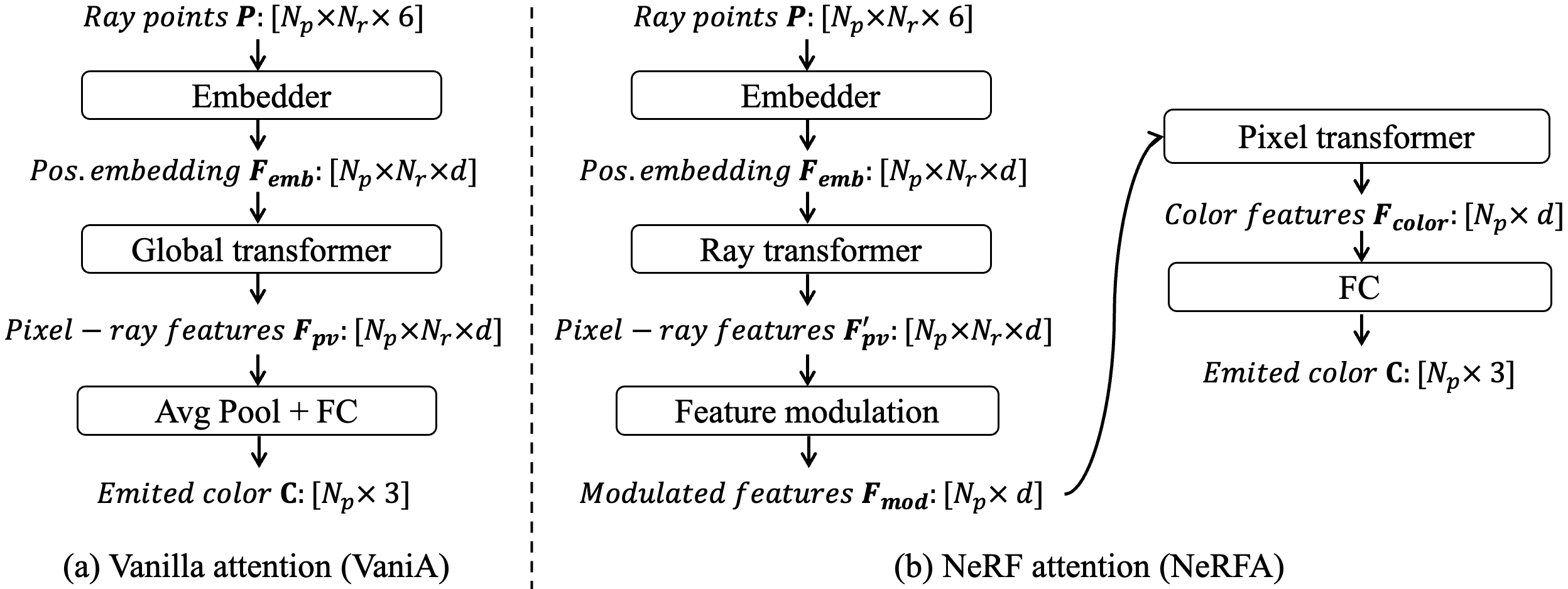}
  \caption{Illustration of the vanilla attention model (Sec.~\ref{sec:vanilla_atten}) and our proposed NeRF attention model (Sec.~\ref{sec:nerfa}). \textbf{Left:} The vanilla attention model simply adopting self-attention, and the pooling and fully connected (FC) layers to get the predicted color results. \textbf{Right:} NeRFA proposes a feature modulation procedure to conduct latent rendering. It uses a ray transformer and a pixel transformer to learn interactions of rays and pixels. A FC layer then summarizes the results and predicts the final color.}
  \label{Fig:nerfa}
\end{figure}

\subsection{Vanilla attention}
\label{sec:vanilla_atten}
Based on the seq2seq framework, we first introduce the vanilla attention (VaniA) solution. It is illustrated on the left side of Figure~\ref{Fig:nerfa}. The first step is to transfer the input ray points $\mathbf{P}$ to positional embedding, essential for view synthesis. The embedder of NeRF~\cite{nerf} is expressed as
\begin{equation}
    \mathbf{F_{emb}}=\mbox{MLP}(\mathbf{P}) \circ \mbox{CosEmbed}(\mathbf{P}),
\label{eq:emb}
\end{equation}
where $\mathbf{F_{emb}}$ denotes the positional embedding matrix with shape $N_p\times N_r \times d$ where $d$ is the hidden dimension size. 

We then send the embedding matrix $\mathbf{F_{emb}}$ to the transformer~\cite{transformer} block to produce pixel-ray features $\mathbf{F_{pv}}\in \mathbb{R}^{N_p \times N_r \times d}$. The transformer block is denoted as 
\begin{equation}
    \mathbf{F_{pv}} = \mbox{GlobalTransformer}(\mathbf{F_{emb}}) = \mbox{SelfAtten}(\mbox{LN}(\mathbf{F_{emb}})) + \mathbf{F_{emb}},
\label{eq:standard_pv}
\end{equation}
where $\mbox{LN}$ is layer normalization and $\mbox{SelfAtten}$ represents the standard self-attention~\cite{transformer} of
\begin{equation}
\mbox{SelfAtten}(\mathbf{X}) = \mbox{softmax}\left(\frac{\mathbf{X} \mathbf{X}^{T}}{\sqrt{d}}\right)\mathbf{X}.
\label{eq:pv}
\end{equation}
The transformer contains $L$-layer transformer blocks and each block contains a $H$-head attention where $L$ and $H$ are two hyper-parameters. We omit the multi-layer and multi-head implementation of transformers for simplicity in all equations. We call this transformer in Eq.~\eqref{eq:standard_pv} ``global transformer" because the sequence length is $N_pN_r$ (that is, the embedding matrix is reshaped to a two-dimensional tensor of shape $N_pN_r \times d$). We accumulate the latent features along with rays via average pooling over the ray dimension, and use fully connected layers (FC) to obtain the final color prediction $\mathbf{C}$ as
\begin{equation}
\mathbf{C}=\mathbf{FC}\left(\frac{1}{N_r}\sum_{i=1}^{N_r}\mathbf{F_{pv}}[:, i, :]\right).
\label{eq:final}
\end{equation}
Our experiments (Sec.~\ref{sec:ablation}) show that vanilla attention does not satisfyingly model the rendering process because the synthesized view misses a lot of high-frequency information. This observation aligns with the recent finding of \cite{nerf, tancik2020fourier}. Besides, the vanilla attention is computational heavy because of the global transformer. The following section proposes the NeRF attention to address these key issues.

\subsection{NeRF attention}
\label{sec:nerfa}
\myparagraph{Positional embedding and ray transformer.} Refer to the right of Fig.~\ref{Fig:nerfa}, our NeRF attention first produces positional embedding via Eq.~\eqref{eq:emb}. After that, the NeRF computes the ray attention to get the pixel-ray features. Different from Eq.~\eqref{eq:pv}, the ray attention is only computed for each ray independently. We write the pixel-ray feature $\mathbf{F'_{pv}}$ as 
\begin{equation}
\mathbf{F'_{pv}}[p] = \mbox{RayTransformer}(\mathbf{F_{emb}}[p]) = \mbox{SelfAtten}(\mbox{LN}(\mathbf{F_{emb}}[p])) + \mathbf{F_{emb}}[p], \forall p \in [1, N_p].
\label{eq:pv_prime}
\end{equation}
The global transformer in vanilla attention (Eq.~\eqref{eq:emb}) has complexity $O(N_p^2 N_r^2d)$ since $N_p$ is much larger than $d$~\cite{deformableDETR}. In contrast, the ray transformer in the NeRF attention (Eq.~\eqref{eq:pv}) reduces the complexity to $O(N_pN_r^2d)$, making training more efficient.

\myparagraph{Feature modulation.} Inspired by NeRF~\cite{nerf}, we modulate the pixel-ray feature to learn volumetric rendering. In NeRF~\cite{nerf}, volumetric rendering is expressed as 
\begin{equation}
C(\mathbf{r}) = \sum_{i=1}^{N_r} \left(e^{-\sum_{j=1}^{i-1}\sigma_j\delta_j}(1-e^{-\sigma_i\delta_i})\mathbf{c_i} \right),
\label{eq:nerf}
\end{equation}
where $C(\mathbf{r})$ is the color corresponding to a traced ray $\mathbf{r}$, $\sigma_j$ denotes the predicted density, $\mathbf{c_i}$ is the predicted color of a ray point, and $\delta_i$ is the distance between adjacent samples. We construct the modulated feature matrix $\mathbf{F_{mod}}\in \mathbb{R}^{N_p \times d}$ from the pixel-ray feature as
\begin{equation}
    \mathbf{F_{mod}}[p]= \sum_{i=1}^{N_r} \left(\mbox{exp}(-\sum_{j=1}^{i-1}\delta_j\mathbf{F'_{pv}}[p][j])\circ\left(\mathbf{1}-\mbox{exp}(-\delta_i \mathbf{F'_{pv}}[p][i])\right)\circ \mathbf{F'_{pv}}[p][i] \right),
\label{eq:mod}
\end{equation}
where the $\mbox{exp}(\mathbf{v})$ is an element-wise exponential function for a vector $\mathbf{v}$ and $\circ$ represents the element-wise product operator. The modulation function in Eq.~\eqref{eq:mod} differs from the NeRF equation (Eq.~\eqref{eq:nerf}) because it does not rely on the explicit prediction of volume density and color and it conducts rendering in the latent space. Our experiments show that the modulated features help model the high-frequency components of the scene.

\myparagraph{Pixel transformer and output.} We adopt a pixel transformer to enable pixel-wise interactions. The pixel transformer takes the modulated features $\mathbf{F_{mod}}$ as input, and models interactions of pixel-level features. This process is formulated as
\begin{equation}
\mathbf{F_{color}}=\mbox{PixelTransformer}(\mathbf{F_{mod}})=\mbox{SelfAtten}(\mbox{LN}(\mathbf{F_{mod}})) + \mathbf{F_{mod}},
\label{eq:pixel_transformer}
\end{equation}
where $\mathbf{F_{color}}\in \mathbb{R}^{N_p \times d}$ is the color embeddings. This operation in Eq.~\eqref{eq:pixel_transformer} yields complexity $O(N_p^2d)$. Finally, we use a fully connected layer to get color $\mathbf{C} = \mbox{FC}(\mathbf{F_{color}})$. We empirically find that the pixel transformer contributes to the fine-grained details in the synthesized image in experiments.

\myparagraph{Overall complexity analysis.} The overall time complexity of the NeRFA model is $O(N_pN_r^2d + N_p^2d)$, which is much lower than the vanilla attention model $O(N_p^2N_r^2d)$. The NeRFA model can be trained within three days on a single NVIDIA V100 GPU for 300k iterations, comparable to NeRF. Note that NeRF training needs one or two days according to statistics in~\cite{nerf}. The speed of NeRFA can be further optimized via recent techniques, such as the fast NeRF~\cite{fastnerf} or the efficient transformer~\cite{kitaev2020reformer}.

%% file: text/4_experiments.tex
\begin{figure}[!t]
  \centering
  \includegraphics[width=\textwidth]{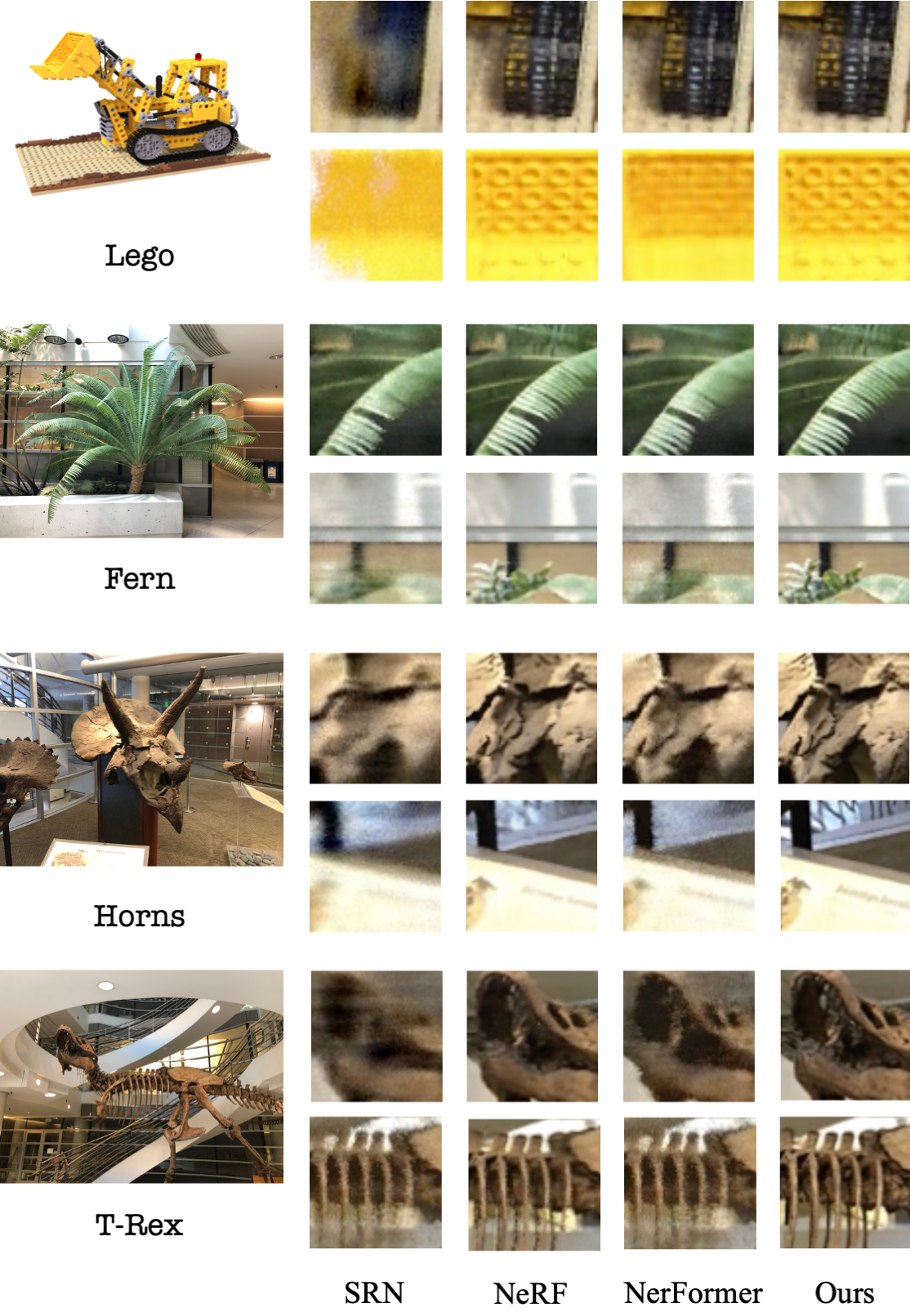}
  \caption{Single-scene view synthesis results on the test split of the Blender~\cite{nerf} and LLFF~\cite{localLightFieldFusion}. Our approach is compared to SRN~\cite{sceneRepresentationNetworks}, NeRF~\cite{nerf}, and NerFormer~\cite{nerformer}. We observe that NerFormer would deteriorate the NeRF but our NeRFA model improves the NeRF and produces sharper details.}
  \label{Fig:vis_result}
\end{figure}

\section{Experiments}
In this section, we compare NeRFA to NeRF~\cite{nerf}, NerFormer~\cite{nerformer} and other baselines on four datasets. We also conduct ablation studies to verify the effectiveness of NeRFA components. A demo video of our approach is provided in the supplementary material.

\subsection{Datasets}
\myparagraph{DeepVoxels.} The DeepVoxels dataset~\cite{deepvoxels_sitzmann} includes four objects: vase, pedestal, chair, and cube. The objects are relatively simple in geometry. The training sets are formed by rendering objects via 479 poses uniformly sampled in the northern hemisphere. We reserve thirty images for validation. Another 1000 testing views are rendered on an Archimedean spiral in the northern hemisphere. All images are of $512\times 512$ resolution.

\myparagraph{Blender.} The Blender dataset is presented in the original NeRF~\cite{nerf} paper. This dataset contains eight objects (materials, drums, etc.) with complicated geometry and non-Lambertian effect. Each object is rendered for 100 training views and 200 testing views. We reserve ten training views for a validation split. The resolution is $800 \times 800$ pixels.

\myparagraph{LLFF.} The LLFF~\cite{localLightFieldFusion} dataset includes eight rendered scenes in total. All are captured by a cellphone with $20-62$ images where $\frac{1}{8}$ of the images are reserved for the test. We reserve another $\frac{1}{8}$ of all images to form the validation split to tune model hyper-parameters. The resolution of the rendered frames is $1008\times 756$.

\begin{table}[!t]
  \caption{Experimental results on the test split of three datasets (DeepVoxels~\cite{deepvoxels_sitzmann}, Blender~\cite{nerf} and LLFF~\cite{localLightFieldFusion}) under the single-scene view synthesis setting. We highlight the best results in \textbf{bold}.}
  \label{tb:res_three_datasets}
  \centering
\resizebox{\linewidth}{!}{
\begin{tabular}{c|ccc|ccc|ccc}
\toprule
 & \multicolumn{3}{c|}{DeepVoxels~\cite{deepvoxels_sitzmann}} & \multicolumn{3}{c|}{Blender~\cite{nerf}} & \multicolumn{3}{c}{LLFF~\cite{localLightFieldFusion}} \\
 & PSNR$\uparrow$ & SSIM$\uparrow$ & LPIPS$\downarrow$ & PSNR$\uparrow$ & SSIM$\uparrow$ & LPIPS$\downarrow$ & PSNR$\uparrow$ & SSIM$\uparrow$ & LPIPS$\downarrow$ \\ \midrule
SRN~\cite{sceneRepresentationNetworks} & 33.15 & 0.947 & 0.069 & 22.15 & 0.843 & 0.164 & 22.78 & 0.654 & 0.369 \\
NV~\cite{neuralVolumes} & 29.57 & 0.930 & 0.094 & 26.31 & 0.879 & 0.158 & - & - & - \\
LLFF~\cite{localLightFieldFusion} & 34.21 & 0.986 & 0.045 & 24.95 & 0.933 & 0.113 & 24.14 & 0.801 & 0.204 \\
NeRF~\cite{nerf} & 40.14 & 0.990 & 0.021 & 32.44 & 0.956 & 0.076 & 26.49 & 0.812 & 0.235 \\
NerFormer~\cite{nerformer} & 35.51 & 0.975 & 0.030 & 28.53 & 0.941 & 0.095 & 25.61 & 0.821 & 0.218 \\
NeRFA (Ours) & \textbf{40.31} & \textbf{0.992} & \textbf{0.020} & \textbf{33.11} & \textbf{0.959} & \textbf{0.072} & \textbf{28.14} & \textbf{0.832} & \textbf{0.198} \\ \bottomrule
\end{tabular}}
\end{table}

\begin{table}[!t]
  \caption{Experimental results on the CO3D~\cite{nerformer} under the category-centric novel view synthesis setting. Baseline numbers are taken from~\cite{nerformer}. The best results are highlighted in \textbf{bold}.}
  \label{tb:co3d}
  \centering
\resizebox{\linewidth}{!}{
\begin{tabular}{c|cccccccc|cccccccc}
\toprule
 & \multicolumn{8}{c|}{$\texttt{Train-unseen}$} & \multicolumn{8}{c}{$\texttt{Test-unseen}$} \\ \cline{2-17} 
 & \multicolumn{2}{c}{\begin{tabular}[c]{@{}c@{}}Average\\ statistics\end{tabular}} & \multicolumn{3}{c}{\begin{tabular}[c]{@{}c@{}}PSNR @\\ \# source views\end{tabular}} & \multicolumn{3}{c|}{\begin{tabular}[c]{@{}c@{}}PSNR @\\ target view difficulty\end{tabular}} & \multicolumn{2}{c}{\begin{tabular}[c]{@{}c@{}}Average\\ statistics\end{tabular}} & \multicolumn{3}{c}{\begin{tabular}[c]{@{}c@{}}PSNR @\\ \# source views\end{tabular}} & \multicolumn{3}{c}{\begin{tabular}[c]{@{}c@{}}PSNR @\\ target view difficulty\end{tabular}} \\
 & PSNR$\uparrow$ & LPIPS$\downarrow$ & 9 & 5 & 1 & easy & med. & hard & PSNR$\uparrow$ & LPIPS$\downarrow$ & 9 & 5 & 1 & easy & med. & hard \\ \midrule
NV-WCE~\cite{neuralVolumes} & 12.3 & 0.34 & 12.5 & 12.3 & 12.0 & 12.4 & 12.0 & 13.6 & 11.6 & 0.35 & 11.7 & 11.6 & 11.6 & 11.7 & 11.2 & 12.0 \\
P3DMesh~\cite{pytorch3d} & 17.2 & \textbf{0.23} & 17.6 & 17.4 & 16.2 & 17.5 & \textbf{16.8} & \textbf{16.0} & 12.4 & 0.26 & 12.6 & 12.5 & 12.1 & 12.6 & 11.8 & 13.2 \\
IPC-WCE~\cite{wiles2020synsin} & 14.1 & 0.36 & 14.4 & 14.2 & 13.4 & 14.4 & 13.7 & 13.4 & 13.8 & 0.27 & 13.8 & 13.7 & 12.6 & 13.8 & 12.8 & 12.2 \\
SRN-WCE~\cite{sceneRepresentationNetworks} & 16.6 & 0.26 & 17.0 & 16.7 & 15.8 & 16.9 & 15.7 & 14.5 & 14.6 & 0.27 & 14.9 & 14.8 & 13.9 & 14.9 & 13.7 & 14.8 \\
NeRF-WCE~\cite{nerf-wce} & 14.3 & 0.27 & 14.3 & 14.9 & 14.2 & 12.1 & 13.6 & 13.6 & 13.8 & 0.27 & 14.3 & 14.9 & 14.2 & 14.4 & 13.0 & 13.0 \\
NerFormer~\cite{nerformer} & 17.9 & 0.26 & 19.3 & 18.3 & 15.6 & 18.9 & 15.5 & 14.6 & 17.6 & 0.27 & 18.9 & 18.1 & 15.1 & 18.6 & 14.9 & 14.7 \\
NeRFA (Ours) & \textbf{18.6} & 0.25 & \textbf{20.2} & \textbf{19.1} & \textbf{16.3} & \textbf{19.7} & 16.1 & 14.9 & \textbf{18.7} & \textbf{0.26} & \textbf{19.7} & \textbf{18.6} & \textbf{15.3} & \textbf{19.4} & \textbf{15.2} & \textbf{14.9} \\ \bottomrule
\end{tabular}}
\end{table}

\myparagraph{CO3D.} The Common Objects in 3D (CO3D~\cite{nerformer}) is a recent dataset containing 1.5 million multi-view images of around 19k objects. This dataset measure the ability of novel view synthesis in the wild. The $18,619$ videos are divided following official release~\cite{nerformer} into four splits: $\texttt{train-known}$, $\texttt{train-unseen}$, $\texttt{test-known}$, and $\texttt{test-unseen}$. The difference between $\texttt{train-unseen}$ and $\texttt{test-unseen}$ is that the $\texttt{train-unseen}$ contains unseen frames in seen videos while $\texttt{test-unseen}$ contains only unseen videos. We focus on the setting of category-centric view synthesis. Under this setting, a single model is trained on the $\texttt{train-known}$ split and is evaluated on 1000 testing samples of $\texttt{train-unseen}$ and $\texttt{test-unseen}$. Evaluation uses $10$ object categories. We do not use the auto-decoders~\cite{nerformer} in all baselines for fairness.

\subsection{Experimental details}
We generate $N_p=128$ rays in a batch for each training iteration, and we sample $N_v=64$ ray points along each ray. The optimizer is Adam~\cite{kingma2014adam} with an initial learning rate of $5\times 10^{-4}$ and an exponential decay of $5\times 10^{-5}$. The hidden dimension of the attention functions is $d=64$, and the head number is $H=8$. On the CO3D~\cite{nerformer} dataset, we use $L=3$ layers of transformers, and on the other datasets, we only use $L=1$. The hyperparameters are the same for all compared approaches. We use three metrics to evaluate NeRFA and the baselines. They are peak signal-to-noise ratio (PSNR), structural similarity index measure (SSIM), and LPIPS~\cite{lpips}.


\subsection{Baselines}
\myparagraph{NV~\cite{neuralVolumes}, LLFF~\cite{localLightFieldFusion} and SRN~\cite{sceneRepresentationNetworks}.} These baselines are proposed earlier than NeRF~\cite{nerf}. The neural volumes (NV) generate novel views by predicting a discretized voxel grid representing RGB$\alpha$ values with $128^3$ samples. LLFF~\cite{localLightFieldFusion} uses a 3D neural network to predict a multi-plane image (MPI) for each training view. LLFF generates novel views via blending and alpha compositing. SRN~\cite{sceneRepresentationNetworks} is a follow-up work of DeepVoxels~\cite{deepvoxels_sitzmann}, representing a scene via an MLP mapping from each coordinate to a feature vector. A recurrent neural network is built upon the feature vector to predict the step size of the ray. The color vector is decoded from the final feature vector. On the CO3D dataset, we use the WCE~\cite{nerf-wce} technique to augment these methods following~\cite{nerformer}.

\myparagraph{NeRF~\cite{nerf}.} NeRF uses an eight-layer MLP with ReLU activations (the hidden dimension is $256$) to learn the volume density. Another FC layer ($128$ channels with a ReLU activation) is added to generate the view-dependent color vector. NeRF adopts the standard volumetric rendering~\cite{optical_models_volume_rendering} to render views based on the MLP predictions. Besides, NeRF introduces a hierarchical volume sampling procedure to avoid sampling occluded regions or free space.

\myparagraph{NerFormer~\cite{nerformer}.} The NerFormer replaces the MLP in NeRF with 3D transformer modules and pooling heads. NerFormer consumes multi-source inputs, uses a series of 3D transformers to learn the ray interaction and uses pooling layers to aggregate per-pixel information. It keeps other components of NeRF, such as the volumetric rendering. In contrast, NeRFA has different input and is of a different model architecture.

\myparagraph{NeRF-WCE~\cite{nerf-wce}} NeRF-WCE augments the NeRF model via warp-conditioned embedding (WCE), which is formed by sampling a tensor of the source image feature. WCE helps capture invariance of rigid scene misalignment. When multiple source views are given, the NeRF-WCE adopts the concatenation of the mean and standard deviation of the view-dependent WCEs as the aggregated WCE.

Besides these baselines, we also compare with other reconstruction-based approaches, P3DMesh~\cite{pytorch3d} and IPC-WCE~\cite{wiles2020synsin}, on the CO3D dataset.

\begin{table}[!t]
  \caption{Ablation study on the validation split of three datasets (DeepVoxels~\cite{deepvoxels_sitzmann}, Blender~\cite{nerf} and LLFF~\cite{localLightFieldFusion}). VaniA, FM, RT, and PT represent the vanilla attention, feature modulation, ray transformer and pixel transformer correspondingly.}
  \label{tb:abl_nerf}
  \centering
\resizebox{\linewidth}{!}{
\begin{tabular}{c|ccc|ccc|ccc}
\toprule
 & \multicolumn{3}{c|}{DeepVoxels~\cite{deepvoxels_sitzmann}} & \multicolumn{3}{c|}{Blender~\cite{nerf}} & \multicolumn{3}{c}{LLFF~\cite{localLightFieldFusion}} \\
 & PSNR$\uparrow$ & SSIM$\uparrow$ & LPIPS$\downarrow$ & PSNR$\uparrow$ & SSIM$\uparrow$ & LPIPS$\downarrow$ & PSNR$\uparrow$ & SSIM$\uparrow$ & LPIPS$\downarrow$ \\ \midrule
NeRFA & 40.30 & 0.991 & 0.024 & 33.17 & 0.941 & 0.074 & 28.15 & 0.833 & 0.192 \\
VaniA & 23.42 & 0.783 & 0.213 & 22.35 & 0.761 & 0.224 & 21.49 & 0.731 & 0.235 \\
w/o FM & 24.51 & 0.791 & 0.194 & 25.14 & 0.793 & 0.195 & 25.42 & 0.801 & 0.208 \\
w/o RT & 31.28 & 0.941 & 0.075 & 28.04 & 0.811 & 0.193 & 26.38 & 0.803 & 0.191 \\
w/o PT & 37.14 & 0.973 & 0.041 & 32.42 & 0.938 & 0.076 & 28.11 & 0.818 & 0.193 \\ \bottomrule
\end{tabular}}
\end{table}

\begin{figure}[!t]
  \centering
  \includegraphics[width=\textwidth]{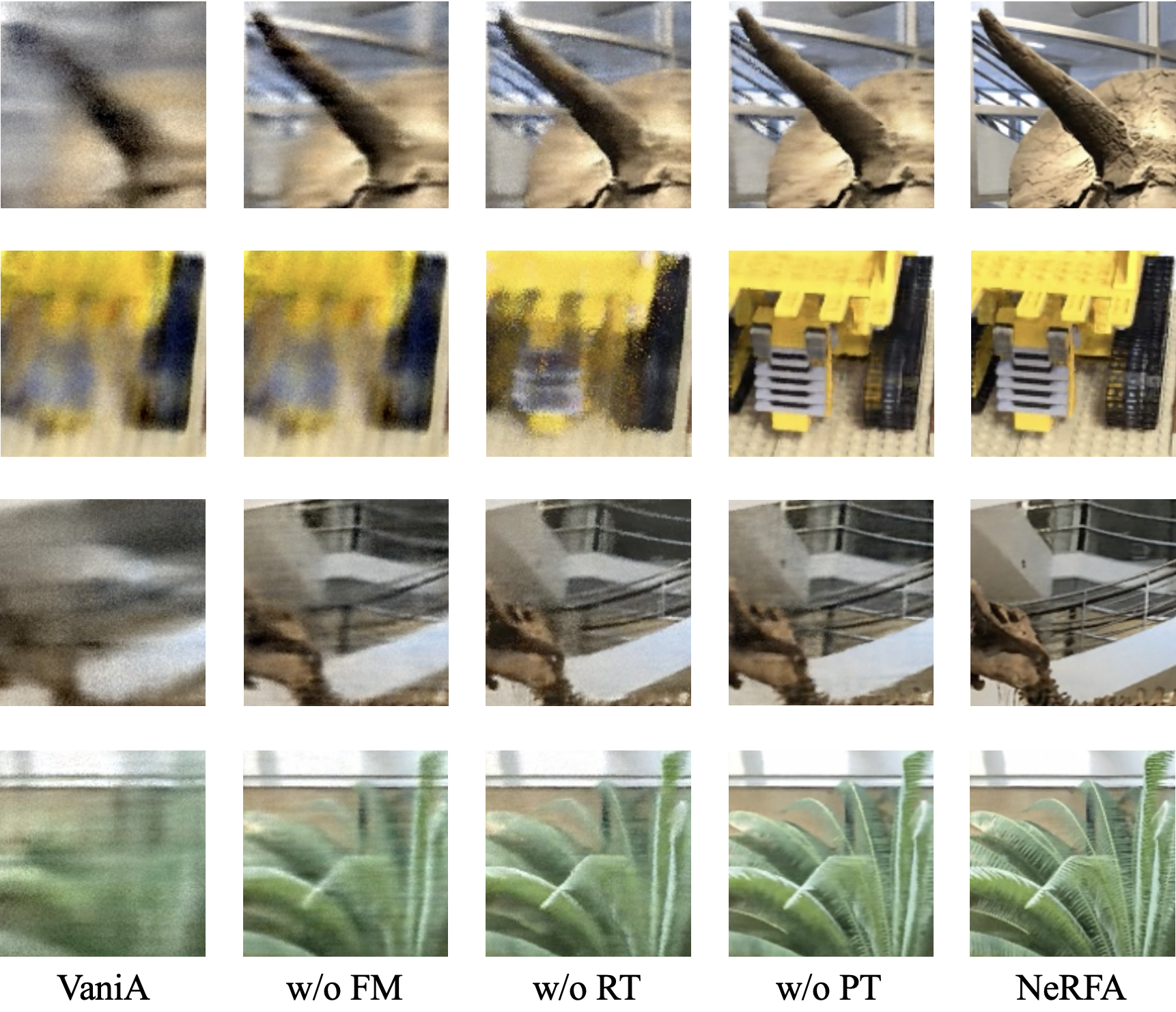}
  \caption{Visualization of the ablated models. We denote the vanilla attention, feature modulation, ray transformer and pixel transformer as VaniA, FM, RT, and PT correspondingly.}
  \label{Fig:ablation_result}
\end{figure}

\subsection{Comparison}
\myparagraph{Single-scene view synthesis.} We first present the results under the single-scene view synthesis setting. This setting is the same as that of NeRF~\cite{nerf}, where one model is trained for each scene. Results are presented in Table~\ref{tb:res_three_datasets} and Figure~\ref{Fig:vis_result}. Under this setting, SRN~\cite{sceneRepresentationNetworks} does not learn enough details and generates blurry results. Although NerFormer~\cite{nerformer} replaces the MLP in NeRF~\cite{nerf} with a group of transformers, the performance is not boosted. In contrast, our NeRFA model successfully improves NeRF by generating sharper details (e.g., the edges of the fern and text of the horns). 

\myparagraph{Category-centric novel-view synthesis.} We report results under the category-centric novel-view synthesis setting. We also test the models with varying numbers of views and difficulty of target views. The numbers are in Table~\ref{tb:co3d}, and some visualizations are provided in the supplementary. The NeRFA model outperforms NerFormer~\cite{nerformer}, NeRF-WCE~\cite{nerf-wce} and other baselines on the $\texttt{test-unseen}$ split. NeRFA is also robust under different numbers of source views and different target-view difficulties. The mesh-based method P3D-mesh~\cite{pytorch3d} yields competitive results on challenging scenarios of the $\texttt{train-unseen}$ split. However, it works poorly on unseen videos in the $\texttt{test-unseen}$ split.

\subsection{Ablation study}
\label{sec:ablation}
In this subsection, we investigate the effects of each NeRFA component. Besides the vanilla attention (VaniA), which is presented in Sec.~\ref{sec:vanilla_atten}, we compare the following ablated variants of NeRFA. The unmentioned model components are the same as the NeRFA model.

\myparagraph{NeRFA w/o feature modulation (w/o FM).} We remove the feature modulation step from the NeRFA model. An average pooling layer replaces the feature modulation layer. The pooled features $\mathbf{F_{pool}}\in \mathbb{R}^{N_p \times d}$ is then sent to the pixel transformer to get the color features $\mathbf{F_{color}}$.

\myparagraph{NeRFA w/o ray transformer (w/o RT).} We remove the ray transformer, and the positional embedding is directly sent to the feature modulation layer. In other words, we set the positional embeddings as the pixel-ray features $\mathbf{F'_{pv}} = \mathbf{F_{emb}}$.

\myparagraph{NeRFA w/o pixel transformer (w/o PT).} In this variant, the pixel transformer is removed from the NeRFA model, and the modulated features are directly sent to the final FC layer. In other words, the color features are set to the modulated features $\mathbf{F_{color}} = \mathbf{F_{mod}}$.

All the ablated models share the same hyper-parameters and training procedures. We present the results on the validation split of three datasets (DeepVoxels~\cite{deepvoxels_sitzmann}, Blender~\cite{nerf} and LLFF~\cite{localLightFieldFusion}) in Table~\ref{tb:abl_nerf}. The visualizations are presented in Figure~\ref{Fig:ablation_result}. We observe that the feature modulation effectively helps learn the high-frequency components of the scene. The ray transformer reduces the blurry and distorted renderings. We also observe that the pixel transformer enhances sharp details.

%% file: text/5_conclusion.tex
\section{Conclusions and future work}
We have presented a simple end-to-end framework for view synthesis. Because vanilla attention could not model the rendering procedure well, we proposed the NeRF attention (NeRFA) model, which considered the NeRF rendering process a soft feature generation procedure. The NeRFA model used ray and pixel transformers to learn connections between rays and pixels. We empirically verified that NeRFA significantly improved vanilla attention and helped learn high-frequency components. NeRFA also demonstrated superior performance over strong baselines such as NerFormer~\cite{nerformer} and NeRF-WCE~\cite{nerf-wce}. Our work suggests multiple future work directions, such as exploring better transformer models~\cite{SwinTransformer} under the seq2seq formulation and learning representations across categories \cite{geoViewSynthesis}. Learning to reconstruct surfaces~\cite{neus} will also be our target.

%% file: neurips_2022.bbl
\begin{thebibliography}{10}
\providecommand{\url}[1]{\texttt{#1}}
\providecommand{\urlprefix}{URL }
\providecommand{\doi}[1]{https://doi.org/#1}

\bibitem{foundationModels}
Bommasani, R., Hudson, D.A., Adeli, E., Altman, R., Arora, S., von Arx, S.,
  Bernstein, M.S., Bohg, J., Bosselut, A., Brunskill, E., et~al.: On the
  opportunities and risks of foundation models. arXiv preprint arXiv:2108.07258
   (2021)

\bibitem{DETR}
Carion, N., Massa, F., Synnaeve, G., Usunier, N., Kirillov, A., Zagoruyko, S.:
  End-to-end object detection with transformers. In: European Conference on
  Computer Vision. pp. 213--229. Springer (2020)

\bibitem{directVoxelOptimization}
Cheng, S., Min, S., Hwann-Tzong, C.: Direct voxel grid optimization: Super-fast
  convergence for radiance fields reconstruction. arXiv preprint
  arXiv:2111.11215  (2021)

\bibitem{bert}
Devlin, J., Chang, M.W., Lee, K., Toutanova, K.: Bert: Pre-training of deep
  bidirectional transformers for language understanding. arXiv preprint
  arXiv:1810.04805  (2018)

\bibitem{visionTransformer}
Dosovitskiy, A., Beyer, L., Kolesnikov, A., Weissenborn, D., Zhai, X.,
  Unterthiner, T., Dehghani, M., Minderer, M., Heigold, G., Gelly, S., et~al.:
  An image is worth 16x16 words: Transformers for image recognition at scale.
  arXiv preprint arXiv:2010.11929  (2020)

\bibitem{vqganTamingTransformer}
Esser, P., Rombach, R., Ommer, B.: Taming transformers for high-resolution
  image synthesis. In: Proceedings of the IEEE/CVF Conference on Computer
  Vision and Pattern Recognition. pp. 12873--12883 (2021)

\bibitem{DeepView}
Flynn, J., Broxton, M., Debevec, P., DuVall, M., Fyffe, G., Overbeck, R.,
  Snavely, N., Tucker, R.: Deepview: View synthesis with learned gradient
  descent. In: Proceedings of the IEEE/CVF Conference on Computer Vision and
  Pattern Recognition. pp. 2367--2376 (2019)

\bibitem{dynamicNeRF}
Gafni, G., Thies, J., Zollhofer, M., Nie{\ss}ner, M.: Dynamic neural radiance
  fields for monocular 4d facial avatar reconstruction. In: Proceedings of the
  IEEE/CVF Conference on Computer Vision and Pattern Recognition. pp.
  8649--8658 (2021)

\bibitem{container}
Gao, P., Lu, J., Li, H., Mottaghi, R., Kembhavi, A.: Container: Context
  aggregation network. CoRR  \textbf{abs/2106.01401} (2021),
  \url{https://arxiv.org/abs/2106.01401}

\bibitem{fastnerf}
Garbin, S.J., Kowalski, M., Johnson, M., Shotton, J., Valentin, J.: Fastnerf:
  High-fidelity neural rendering at 200fps. https://arxiv.org/abs/2103.10380
  (2021)

\bibitem{Lumigraph}
Gortler, S.J., Grzeszczuk, R., Szeliski, R., Cohen, M.F.: The lumigraph. In:
  Proceedings of the 23rd annual conference on Computer graphics and
  interactive techniques. pp. 43--54 (1996)

\bibitem{nerf-wce}
Henzler, P., Reizenstein, J., Labatut, P., Shapovalov, R., Ritschel, T.,
  Vedaldi, A., Novotny, D.: Unsupervised learning of 3d object categories from
  videos in the wild. In: Proceedings of the IEEE/CVF Conference on Computer
  Vision and Pattern Recognition (CVPR). pp. 4700--4709 (June 2021)

\bibitem{LearningMultiViewStereo}
Kar, A., H\"{a}ne, C., Malik, J.: Learning a multi-view stereo machine. In:
  Guyon, I., Luxburg, U.V., Bengio, S., Wallach, H., Fergus, R., Vishwanathan,
  S., Garnett, R. (eds.) Advances in Neural Information Processing Systems.
  vol.~30. Curran Associates, Inc. (2017),
  \url{https://proceedings.neurips.cc/paper/2017/file/9c838d2e45b2ad1094d42f4ef36764f6-Paper.pdf}

\bibitem{kingma2014adam}
Kingma, D.P., Ba, J.: Adam: A method for stochastic optimization. arXiv
  preprint arXiv:1412.6980  (2014)

\bibitem{kitaev2020reformer}
Kitaev, N., Kaiser, {\L}., Levskaya, A.: Reformer: The efficient transformer.
  arXiv preprint arXiv:2001.04451  (2020)

\bibitem{viewformer}
Kulh{\'a}nek, J., Derner, E., Sattler, T., Babu{\v{s}}ka, R.: Viewformer:
  Nerf-free neural rendering from few images using transformers. arXiv preprint
  arXiv:2203.10157  (2022)

\bibitem{autoint}
Lindell, D., Martel, J., Wetzstein, G.: {AutoInt}: Automatic integration for
  fast neural volume rendering. https://arxiv.org/abs/2012.01714  (2020)

\bibitem{NSVF}
Liu, L., Gu, J., Zaw~Lin, K., Chua, T.S., Theobalt, C.: Neural sparse voxel
  fields. Advances in Neural Information Processing Systems  \textbf{33},
  15651--15663 (2020)

\bibitem{SwinTransformer}
Liu, Z., Lin, Y., Cao, Y., Hu, H., Wei, Y., Zhang, Z., Lin, S., Guo, B.: Swin
  transformer: Hierarchical vision transformer using shifted windows. arXiv
  preprint arXiv:2103.14030  (2021)

\bibitem{neuralVolumes}
Lombardi, S., Simon, T., Saragih, J., Schwartz, G., Lehrmann, A., Sheikh, Y.:
  Neural volumes: Learning dynamic renderable volumes from images. arXiv
  preprint arXiv:1906.07751  (2019)

\bibitem{optical_models_volume_rendering}
Max, N.: Optical models for direct volume rendering. IEEE Transactions on
  Visualization and Computer Graphics  \textbf{1}(2),  99--108 (1995)

\bibitem{localLightFieldFusion}
Mildenhall, B., Srinivasan, P.P., Ortiz-Cayon, R., Kalantari, N.K.,
  Ramamoorthi, R., Ng, R., Kar, A.: Local light field fusion: Practical view
  synthesis with prescriptive sampling guidelines. ACM Transactions on Graphics
  (TOG)  \textbf{38}(4),  1--14 (2019)

\bibitem{nerf}
Mildenhall, B., Srinivasan, P.P., Tancik, M., Barron, J.T., Ramamoorthi, R.,
  Ng, R.: Nerf: Representing scenes as neural radiance fields for view
  synthesis. In: European conference on computer vision. pp. 405--421. Springer
  (2020)

\bibitem{sentimentAnalysis}
Mohammad, S.M.: Sentiment analysis: Detecting valence, emotions, and other
  affectual states from text. In: Emotion measurement, pp. 201--237. Elsevier
  (2016)

\bibitem{rendernet}
Nguyen-Phuoc, T.H., Li, C., Balaban, S., Yang, Y.: Rendernet: A deep
  convolutional network for differentiable rendering from 3d shapes. Advances
  in Neural Information Processing Systems  \textbf{31} (2018)

\bibitem{giraffe}
Niemeyer, M., Geiger, A.: Giraffe: Representing scenes as compositional
  generative neural feature fields. In: Proceedings of the IEEE/CVF Conference
  on Computer Vision and Pattern Recognition. pp. 11453--11464 (2021)

\bibitem{nerfies}
Park, K., Sinha, U., Barron, J.T., Bouaziz, S., Goldman, D.B., Seitz, S.M.,
  Martin-Brualla, R.: Nerfies: Deformable neural radiance fields. ICCV  (2021)

\bibitem{howDoViTWork}
Park, N., Kim, S.: How do vision transformers work? arXiv preprint
  arXiv:2202.06709  (2022)

\bibitem{soft3dreconstruction}
Penner, E., Zhang, L.: Soft 3d reconstruction for view synthesis. ACM
  Transactions on Graphics (TOG)  \textbf{36}(6),  1--11 (2017)

\bibitem{dnerf}
Pumarola, A., Corona, E., Pons-Moll, G., Moreno-Noguer, F.: D-nerf: Neural
  radiance fields for dynamic scenes. In: Proceedings of the IEEE/CVF
  Conference on Computer Vision and Pattern Recognition. pp. 10318--10327
  (2021)

\bibitem{languageMultiTask}
Radford, A., Wu, J., Child, R., Luan, D., Amodei, D., Sutskever, I.: Language
  models are unsupervised multitask learners. OpenAI blog  \textbf{1}(8), ~9
  (2019)

\bibitem{pva}
Raj, A., Zollhoefer, M., Simon, T., Saragih, J., Saito, S., Hays, J., Lombardi,
  S.: Pva: Pixel-aligned volumetric avatars. arXiv preprint arXiv:2101.02697
  (2021)

\bibitem{pytorch3d}
Ravi, N., Reizenstein, J., Novotny, D., Gordon, T., Lo, W.Y., Johnson, J.,
  Gkioxari, G.: Accelerating 3d deep learning with pytorch3d. arXiv preprint
  arXiv:2007.08501  (2020)

\bibitem{kilonerf}
Reiser, C., Peng, S., Liao, Y., Geiger, A.: Kilonerf: Speeding up neural
  radiance fields with thousands of tiny mlps (2021)

\bibitem{nerformer}
Reizenstein, J., Shapovalov, R., Henzler, P., Sbordone, L., Labatut, P.,
  Novotny, D.: Common objects in 3d: Large-scale learning and evaluation of
  real-life 3d category reconstruction. In: Proceedings of the IEEE/CVF
  International Conference on Computer Vision. pp. 10901--10911 (2021)

\bibitem{geoViewSynthesis}
Rombach, R., Esser, P., Ommer, B.: Geometry-free view synthesis: Transformers
  and no 3d priors. In: Proceedings of the IEEE/CVF International Conference on
  Computer Vision (ICCV). pp. 14356--14366 (October 2021)

\bibitem{graf}
Schwarz, K., Liao, Y., Niemeyer, M., Geiger, A.: Graf: Generative radiance
  fields for 3d-aware image synthesis. Advances in Neural Information
  Processing Systems  \textbf{33},  20154--20166 (2020)

\bibitem{texturedNeuralAvatars}
Shysheya, A., Zakharov, E., Aliev, K.A., Bashirov, R., Burkov, E., Iskakov, K.,
  Ivakhnenko, A., Malkov, Y., Pasechnik, I., Ulyanov, D., et~al.: Textured
  neural avatars. In: Proceedings of the IEEE/CVF Conference on Computer Vision
  and Pattern Recognition. pp. 2387--2397 (2019)

\bibitem{deepvoxels_sitzmann}
Sitzmann, V., Thies, J., Heide, F., Niessner, M., Wetzstein, G., Zollhofer, M.:
  Deepvoxels: Learning persistent 3d feature embeddings. In: Proceedings of the
  IEEE/CVF Conference on Computer Vision and Pattern Recognition (CVPR) (June
  2019)

\bibitem{sceneRepresentationNetworks}
Sitzmann, V., Zollh{\"o}fer, M., Wetzstein, G.: Scene representation networks:
  Continuous 3d-structure-aware neural scene representations. Advances in
  Neural Information Processing Systems  \textbf{32} (2019)

\bibitem{PushingBoundariesOfView}
Srinivasan, P.P., Tucker, R., Barron, J.T., Ramamoorthi, R., Ng, R., Snavely,
  N.: Pushing the boundaries of view extrapolation with multiplane images. In:
  Proceedings of the IEEE/CVF Conference on Computer Vision and Pattern
  Recognition (CVPR) (June 2019)

\bibitem{sun2019videobert}
Sun, C., Myers, A., Vondrick, C., Murphy, K., Schmid, C.: Videobert: A joint
  model for video and language representation learning. In: Proceedings of the
  IEEE/CVF International Conference on Computer Vision. pp. 7464--7473 (2019)

\bibitem{lxmert}
Tan, H., Bansal, M.: Lxmert: Learning cross-modality encoder representations
  from transformers. In: Proceedings of the 2019 Conference on Empirical
  Methods in Natural Language Processing and the 9th International Joint
  Conference on Natural Language Processing (EMNLP-IJCNLP). pp. 5103--5114
  (2019)

\bibitem{tancik2020fourier}
Tancik, M., Srinivasan, P., Mildenhall, B., Fridovich-Keil, S., Raghavan, N.,
  Singhal, U., Ramamoorthi, R., Barron, J., Ng, R.: Fourier features let
  networks learn high frequency functions in low dimensional domains. Advances
  in Neural Information Processing Systems  \textbf{33},  7537--7547 (2020)

\bibitem{deferredNeuralRendering}
Thies, J., Zollh{\"o}fer, M., Nie{\ss}ner, M.: Deferred neural rendering: Image
  synthesis using neural textures. ACM Transactions on Graphics (TOG)
  \textbf{38}(4),  1--12 (2019)

\bibitem{MultiViewSupervision}
Tulsiani, S., Zhou, T., Efros, A.A., Malik, J.: Multi-view supervision for
  single-view reconstruction via differentiable ray consistency. In:
  Proceedings of the IEEE conference on computer vision and pattern
  recognition. pp. 2626--2634 (2017)

\bibitem{transformer}
Vaswani, A., Shazeer, N., Parmar, N., Uszkoreit, J., Jones, L., Gomez, A.N.,
  Kaiser, L., Polosukhin, I.: Attention is all you need. arXiv preprint
  arXiv:1706.03762  (2017)

\bibitem{LargeScaleTexturing}
Waechter, M., Moehrle, N., Goesele, M.: Let there be color! large-scale
  texturing of 3d reconstructions. In: European conference on computer vision.
  pp. 836--850. Springer (2014)

\bibitem{neus}
Wang, P., Liu, L., Liu, Y., Theobalt, C., Komura, T., Wang, W.: Neus: Learning
  neural implicit surfaces by volume rendering for multi-view reconstruction.
  NeurIPS  (2021)

\bibitem{ibrnet}
Wang, Q., Wang, Z., Genova, K., Srinivasan, P.P., Zhou, H., Barron, J.T.,
  Martin-Brualla, R., Snavely, N., Funkhouser, T.: Ibrnet: Learning multi-view
  image-based rendering. In: Proceedings of the IEEE/CVF Conference on Computer
  Vision and Pattern Recognition. pp. 4690--4699 (2021)

\bibitem{wiles2020synsin}
Wiles, O., Gkioxari, G., Szeliski, R., Johnson, J.: Synsin: End-to-end view
  synthesis from a single image. In: Proceedings of the IEEE/CVF Conference on
  Computer Vision and Pattern Recognition. pp. 7467--7477 (2020)

\bibitem{SurfaceLight}
Wood, D.N., Azuma, D.I., Aldinger, K., Curless, B., Duchamp, T., Salesin, D.H.,
  Stuetzle, W.: Surface light fields for 3d photography. In: Proceedings of the
  27th annual conference on Computer graphics and interactive techniques. pp.
  287--296 (2000)

\bibitem{xlnet}
Yang, Z., Dai, Z., Yang, Y., Carbonell, J., Salakhutdinov, R., Le, Q.V.: Xlnet:
  Generalized autoregressive pretraining for language understanding. arXiv
  preprint arXiv:1906.08237  (2019)

\bibitem{VolumeRO}
Yariv, L., Gu, J., Kasten, Y., Lipman, Y.: Volume rendering of neural implicit
  surfaces. In: NeurIPS (2021)

\bibitem{REST}
Zhang, Q., Yang, Y.: Rest: An efficient transformer for visual recognition.
  CoRR  \textbf{abs/2105.13677} (2021), \url{https://arxiv.org/abs/2105.13677}

\bibitem{lpips}
Zhang, R., Isola, P., Efros, A.A., Shechtman, E., Wang, O.: The unreasonable
  effectiveness of deep features as a perceptual metric. In: Proceedings of the
  IEEE conference on computer vision and pattern recognition. pp. 586--595
  (2018)

\bibitem{rapl}
Zhao, Z., Gan, C., Wu, J., Guo, X., Tenenbaum, J.: Augmenting policy learning
  with routines discovered from a demonstration. arXiv preprint
  arXiv:2012.12469  (2020)

\bibitem{6dpose}
Zhao, Z., Peng, G., Wang, H., Fang, H.S., Li, C., Lu, C.: Estimating 6d pose
  from localizing designated surface keypoints. arXiv preprint arXiv:1812.01387
   (2018)

\bibitem{ProTo}
Zhao, Z., Samel, K., Chen, B., song, l.: Proto: Program-guided transformer for
  program-guided tasks. In: Ranzato, M., Beygelzimer, A., Dauphin, Y., Liang,
  P., Vaughan, J.W. (eds.) Advances in Neural Information Processing Systems.
  vol.~34, pp. 17021--17036. Curran Associates, Inc. (2021),
  \url{https://proceedings.neurips.cc/paper/2021/file/8d34201a5b85900908db6cae92723617-Paper.pdf}

\bibitem{p3aformer}
Zhao, Z., Wu, Z., Zhuang, Y., Li, B., Jia, J.: Tracking objects as pixel-wise
  distributions (2022). \doi{10.48550/ARXIV.2207.05518},
  \url{https://arxiv.org/abs/2207.05518}

\bibitem{viewSynthesis}
Zhou, T., Tulsiani, S., Sun, W., Malik, J., Efros, A.A.: View synthesis by
  appearance flow. In: European conference on computer vision. pp. 286--301.
  Springer (2016)

\bibitem{von}
Zhu, J.Y., Zhang, Z., Zhang, C., Wu, J., Torralba, A., Tenenbaum, J., Freeman,
  B.: Visual object networks: Image generation with disentangled 3d
  representations. Advances in neural information processing systems
  \textbf{31} (2018)

\bibitem{deformableDETR}
Zhu, X., Su, W., Lu, L., Li, B., Wang, X., Dai, J.: Deformable detr: Deformable
  transformers for end-to-end object detection. arXiv preprint arXiv:2010.04159
   (2020)

\end{thebibliography}
